# Mega-cities dominate China's urban greening

Xiaoxin Zhang[1,*], Martin Brandt[1], Xiaoye Tong[1], Xiaowei Tong[2], Wenmin Zhang[1], Florian Reiner[1], Sizhuo Li[1], Feng Tian[3], Yuemin Yue[2,4], Weiqi Zhou[5], Bin Chen[6], Xiangming Xiao[7], Rasmus Fensholt[1]

[1]Department of Geosciences and Natural Resource Management, University of Copenhagen, Copenhagen, Denmark

[2]Key Laboratory for Agro-ecological Processes in Subtropical Region, Institute of Subtropical Agriculture, Chinese Academy of Sciences, Changsha, China

[3]Hubei Key Laboratory of Quantitative Remote Sensing of Land and Atmosphere, School of Remote Sensing and Information Engineering, Wuhan University, Wuhan, China

[4]Huanjiang Observation and Research Station for Karst Ecosystem, Chinese Academy of Sciences, Huanjiang, China

[5]State Key Laboratory of Urban and Regional Ecology, Research Center for Eco-Environmental Sciences, Chinese Academy of Sciences, Beijing 100085, China

[6]Future Urbanity & Sustainable Environment (FUSE) Lab, Division of Landscape Architecture, Faculty of Architecture, The University of Hong Kong, Hong Kong Special Administrative Region

[7]Department of Microbiology and Plant Biology, Center for Earth Observation and Modeling, University of Oklahoma, Norman, OK, USA

**Corresponding author:**

*To whom correspondence should be addressed. Email: xzh@ign.ku.dk

## Abstract

Trees play a crucial role in urban environments, offering various ecosystem services that contribute to public health and human well-being. China has initiated a range of urban greening policies over the past decades, however, monitoring their impact on urban tree dynamics at a national scale has proven challenging. In this study, we deployed nano-satellites to quantify urban tree coverage in all major Chinese cities larger than 50 km$^2$ in 2010 and 2019. Our findings indicate that approximately 6000 km$^2$ (11%) of urban areas were covered by trees in 2019, and 76% of these cities experienced an increase in tree cover compared to 2010. Notably, the increase in tree cover in mega-cities such as Beijing, and Shanghai was approximately twice as large as in most other cities (7.69% vs 3.94%). The study employs a data-driven approach towards assessing urban tree cover changes in relation to greening policies, showing clear signs of tree cover increases but also suggesting an uneven implementation primarily benefiting a few mega-cities.

# Main

China's rapid urbanization and rural outmigration in the past two decades have led to the creation of millions of new houses and extensive impervious surfaces, often at the expense of agricultural land and forests[1,2]. Mega-cities are often associated with a diminished quality of life due to pervasive environmental issues such as traffic congestion, air pollution and the dominance of concrete landscapes. To improve the well-being of people living in Chinese cities, urban greening policies have been implemented since 1992[3,4]. Trees play a vital role in urban environments, being placed in parks, yards, gardens, and along streets, serving as an essential element of urban life[6,7]. Several studies have documented that urban trees provide benefits for municipalities and their residents, and local, regional, and global initiatives have promoted the planting and preservation of urban trees[6,8]. Urban trees have been reported to mitigate urban heat islands[9,10], reduce energy consumption[11], filter air and water pollution[12], reduce rainfall runoff[13], sequester atmospheric carbon dioxide[14], enhance biodiversity[15], increase property values[16], and improve physical and mental health of residents[17].

Recent studies observed a considerable greening of urban areas in China over the past decade using time-series of satellite images[2,18]. However, it remains unclear to which extent this urban greening is caused by trees, green spaces consisting of grasses and trees as a result of urban greening policies, or if the greening is caused by an increased vegetation growth related primarily to climatic factors[19], such as increasing air temperatures from urban heat islands, elevated atmospheric $CO_2$ concentrations, or nitrogen deposition[20]. This is because cities represent complex and heterogeneous landscapes where vegetation appears in a patchy structure[21]. Trees are often relatively small in size, and a heterogeneous background including green grasses and shadows from tall buildings often causes a mixed pixel signal in freely available satellite imagery having a spatial resolution > 10 m, making it challenging to identify tree canopies[22,23]. Consequently, it is not well known to what extent urban greening policies have been successfully implemented in China towards increasing urban tree cover, and how improvements are balanced between cities at the national scale in relation to environmental conditions and urban development.

The growing availability of sub-meter resolution images from aerial campaigns or commercial satellites, such as WorldView or Gaofen-2, as well as Lidar data enables monitoring of urban trees[22,24,24–26], but these images are costly and typically not available at repeated time steps at city or national scale[22]. This limits their applicability for large-scale urban tree mapping, and only a few countries have nationwide inventories of urban trees[27,28], which are however snapshots in time. The advent of images from the PlanetScope nano-satellite constellation, which provide daily global imagery at a resolution high enough to identify single trees (~3 m), represents an emerging alternative for such large scale mapping. It has been shown that these images can support mapping of individual trees at continental scales[29], but the short period of data availability (since 2017) makes it unfit to study changes over longer time periods. Here, we complement the PlanetScope satellite constellation with data from RapidEye (~5 m), providing a comparable product since 2010, and uncover how China's urban greening policies have been implemented in regards to changes in tree cover across all major cities of China between 2010 and 2019.

## Results

**Uneven distribution of urban tree cover across China's cities**

We used 3-m resolution PlanetScope satellite imagery from 2019[29] covering all Chinese cities with an urban area larger than 50 km$^2$ (242 cities; see Methods for definition of urban areas), summing up to a total area of 51,882 km$^2$. We trained a deep learning segmentation model[30] with labels corresponding to an area of 209 km$^2$ (Supplementary Table 1, Supplementary Fig. 1a) and mapped urban tree canopies, including trees along roads, in parks and in private gardens (Fig. 1) at a level of detail that was previously only possible for single-city surveys using sub-meter resolution imagery[22,26,31] or LIDAR[28]. Our map can capture single trees and small tree clusters classified as "built-up" areas in contemporary land cover products[32] (Fig.1c, Supplementary Fig. 2, Supplementary Table 2). We find that 41.94% of the mapped trees and tree-canopy clusters were smaller than 100 m$^2$ (Supplementary Fig. 3), which is likely to be missed by using satellite imagery with a resolution ≥10 m (Supplementary Fig. 2).

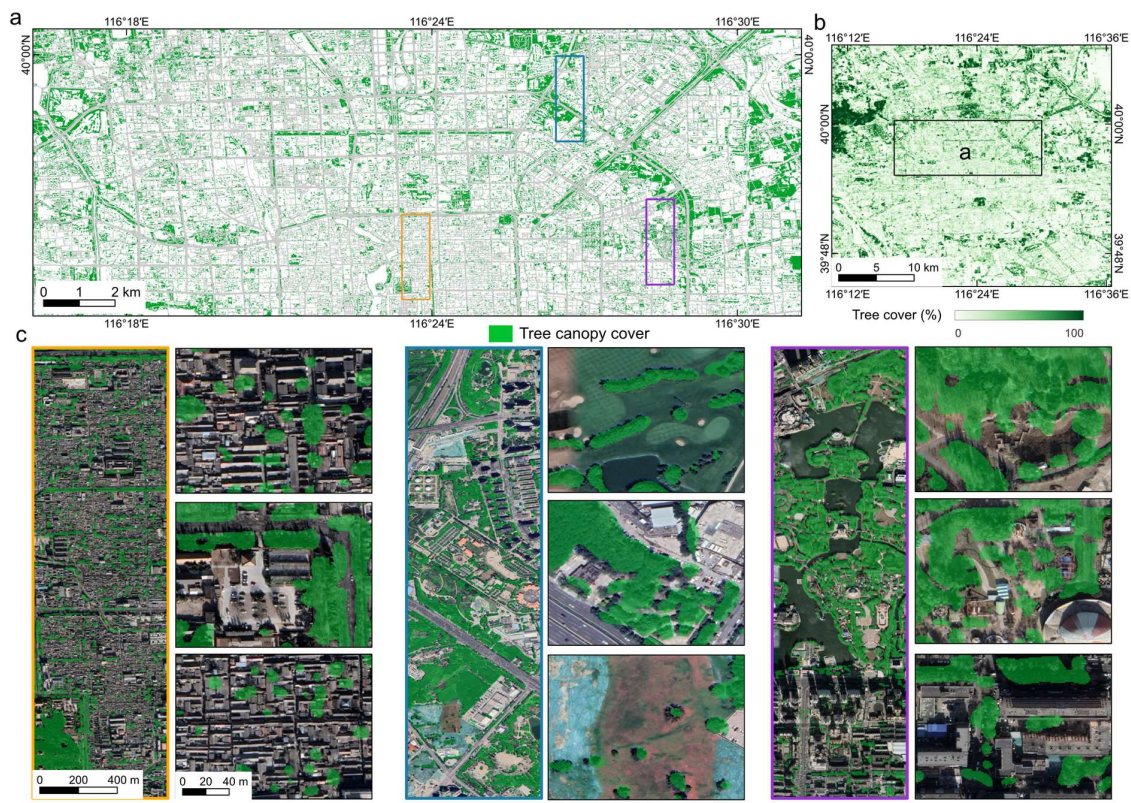

**Fig. 1: Mapping urban tree canopies in China using PlanetScope imagery from 2019**. **a**, Trees canopy cover are mapped for Beijing. **b**, Tree cover in Beijing aggregated to 1 ha (100 m × 100 m). **c**, Zoom-in of PlanetScope-based urban trees overlaid on Google Earth Satellite.

At the city level, the mean urban tree cover in the 242 largest cities of China is 11.47% ($R^2$=0.90, bias=0.37%), and a total area of 5951 km$^2$ is covered by urban trees (Fig. 2, Table 1). Note that the FAO definition (FRA, 2020) for a forest is 10% tree cover per 0.5 ha, implying that 22.45% of China's cities would qualify as a forest if the land use is disregarded. Urban trees are also not evenly distributed among cities. Spatially, the urban tree cover varies across China, with cities in the south-central (12.93%), northeast (12.68%), and southwest (16.70%) having a higher urban tree cover than the national average (Supplementary Fig. 4). On the contrary, cities in the northwestern China characterized by dry climatic conditions show the lowest urban tree cover (6.25%) (Supplementary Fig. 4, 5). Notable examples of cities with lowest urban tree cover are Xilingol (2.14%), Yulin (2.16%), and Aksu (2.18%) (Fig. 2 Supplementary Fig. 4).

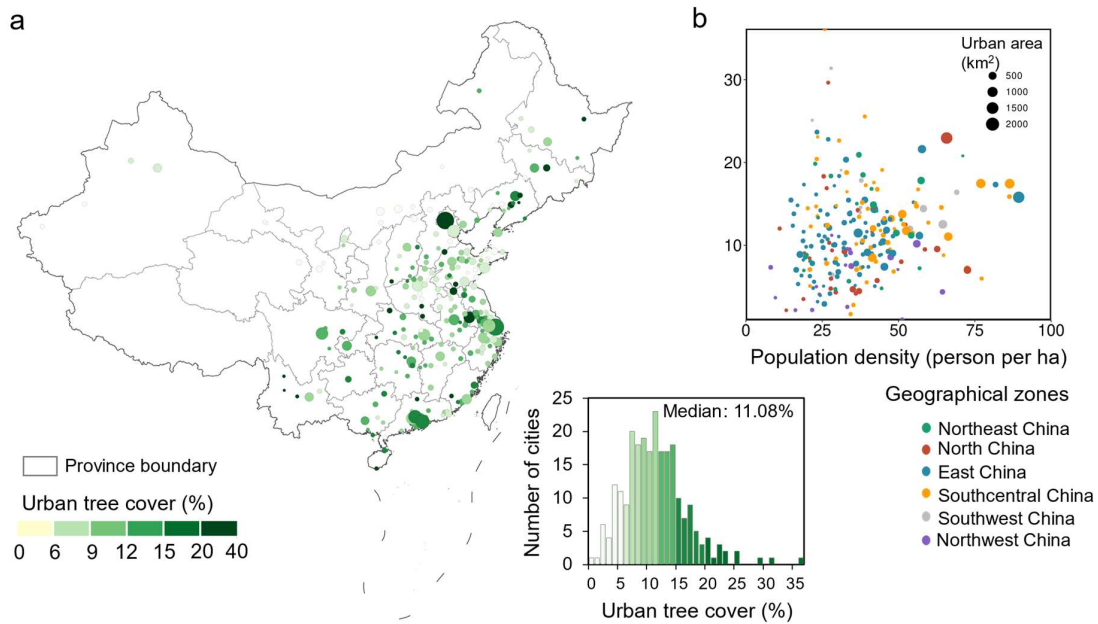

**Fig. 2: Urban tree cover at the city level in 2019**. **a**, Urban tree cover for 242 cities, with each city being represented by a circle and the size of the circle denotes the total tree cover area (frequency plot of urban tree cover inserted). **b**, Urban tree cover with mean population density for 242 cities, divided into geographical zones.

Large cities with dense populations tend to have a higher mean tree cover as compared to small cities (Table 1, Fig. 2a,b, Supplementary Fig. 4b): The two mega-cities Beijing (22.96%) and Shanghai (14.35%) on average have 19.38% tree cover, and the four large cities Shenzhen (17.44%), Guangzhou (17.56%), Suzhou (8.79%), Foshan (11.75%) on average have 14.53% urban tree cover (Table 1, Supplementary Fig. 4b). The 19 medium-sized cities on average have 11.96% tree cover, and the 121 small cities have an average tree cover of 10.89% (Table 1). The 96 emerging cities with urban areas < 100 km² show the highest intra-class variability in urban tree cover, with some cities exhibiting the highest and lowest levels of urban tree cover (Supplementary Fig. 4b). The uneven distribution of urban tree cover is consistent with the size of urban areas. Larges cities, typically regarded as developed cities, tend to have a higher urban tree cover in comparison to less developed cities, or developing cities.

**Table 1. Urban tree cover and population density in 2019 and tree cover changes between 2010 and 2019 grouped by urban area size**. Note that the change includes fewer cities due to the lower number of high-quality RapidEye images available in 2010.

| Unit: km² | | | 2019 | | | Change 2010-2019 | | |
|---|---|---|---|---|---|---|---|---|
| | | Cities | Urban area (km²) | Population density 2019 (person/ha) | Tree cover (%) (bias=0.37%) | Cities | Urban area (km²) | Tree cover (%) (bias=1.07%) |
| (50-100) | Emerging city | 96 | 7019 | 4597 | 11.80 | 55 | 4012 | 3.94 |
| (100-500) | Small city | 121 | 22714 | 5435 | 10.89 | 69 | 13639 | 3.99 |
| (500-1000) | Medium city | 19 | 13115 | 7001 | 11.96 | 16 | 11581 | 3.65 |
| (1000-1500) | Large city | 4 | 4619 | 8013 | 14.53 | 3 | 3596 | 6.09 |
| ≥1500 | Mega-city | 2 | 4077 | 10986 | 19.38 | 2 | 4078 | 7.69 |
| | China | 242 | 51544 | 6389 | 11.47 | 145 | 36906 | 4.57% |

**Changes in urban tree cover between 2010 and 2019**

We acquired high-quality RapidEye satellite imagery for 145 representative major cities and mapped urban trees for 2010 (the earliest phase of the lifetime of the satellite constellation) using the same deep learning framework as applied for the PlanetScope images (see Methods). RapidEye provides an image quality comparable to PlanetScope (Supplementary Fig. 6), but at a less frequent revisiting time, implying that not all the cities analyzed in 2019 could be covered (see Methods, Supplementary Fig. 7).

At the national scale, urban tree cover increased in 76% of the cities from 7.25% ($R^2$=0.84, bias=-0.99%) in 2010 to 11.82 ($R^2$=0.90, bias=0.37%) in 2019; an increase of 4.57% ($R^2$=0.83, bias=1.07%) (Fig. 3a, Supplementary Fig. 8). As also observed for urban tree cover in 2019, the changes in tree cover are not homogeneous across cities and are related to the city size, with the two mega-cities Shanghai (+8.30%) and Beijing (+7.42%) having the highest increase (Fig. 3a, Table 1). Also the class of large cities showed considerable increases in urban tree cover (on average 6%; bias=1.07%) (Table 1). The changes in the remaining three classes of city sizes were on average much lower (3.78%; bias=1.07%), with a high variability for the class covering the smallest cities (Supplementary Fig. 4d). Urban tree cover decreased in 24% of all cities, for example in Chongqing (-6.29%), Hangzhou (-4.73%), and Wuhan (-2.03%) (Fig. 3c).

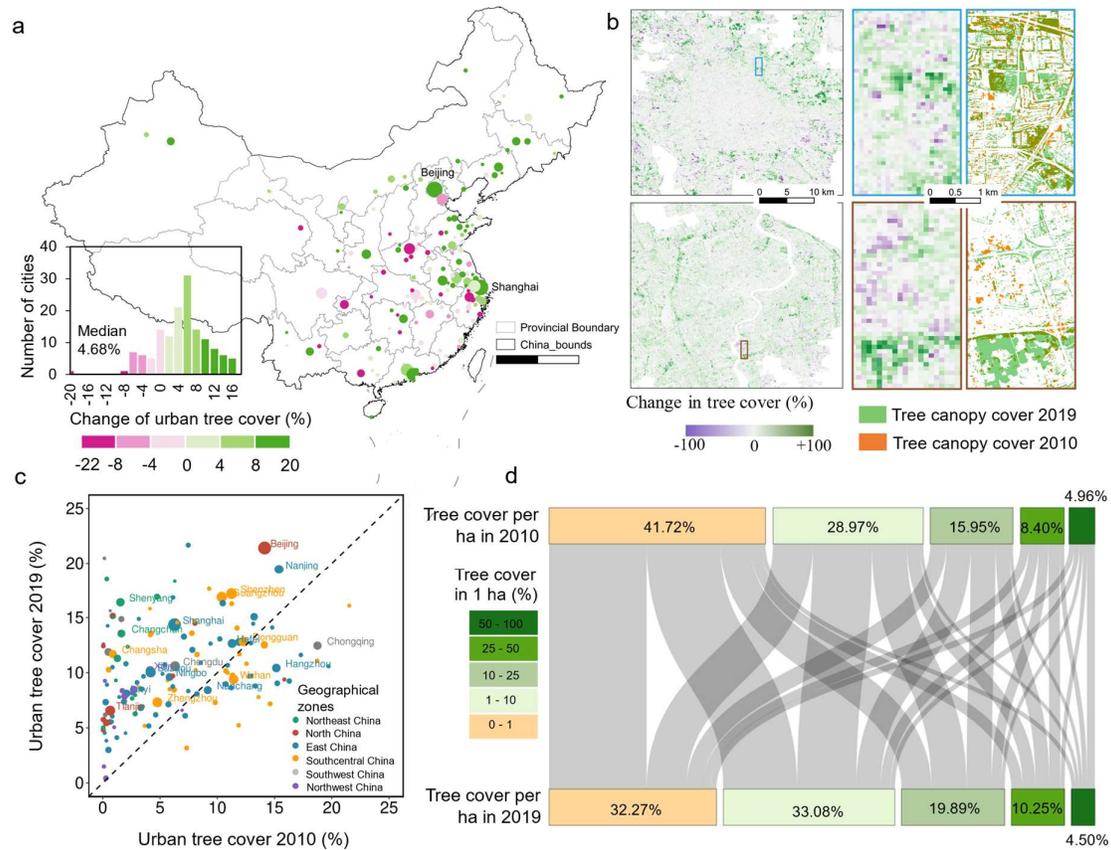

**Fig. 3: Changes in urban tree cover between 2010 and 2019**. **a**, Spatial patterns and frequency plot of changes in urban tree cover at the city level (n = 145). **b**, Change in tree cover for Beijing and Shanghai in 1-ha grids and example patches of tree canopy cover in 2019 and 2010. **c**, Urban tree cover (%) for 2010 (x-axis) and 2019 (y-axis) divided into geographical zones (n = 145). Cities with urban areas ⩾ 500 km$^2$ were labeled with city names. Cities above the 1:1 line experienced an increase in tree cover. **d**, Transitions of tree cover (grouped into intervals of % cover per ha) from 2010 to 2019 for 1-ha grids in urban areas (n = 3,531,113).

We further compared the transition of urban tree cover from 2010 to 2019 grouped into classes of % tree cover for 1-ha grids (Fig. 3d). The 1-ha grids classified as high tree cover (50-100%) observed a slight decrease of 0.45%, probably reflecting suburb forests that were replaced by impervious surfaces. In contrast, grids with no or very low tree cover in 2010 (0-1%) decreased by 9.45%, possibly reflecting the impact of greening policies. Grids with tree cover of 1-10% increased by 4.10%, grids with 10-25% tree cover increased by 3.95%, and grids with tree cover 25-50% increased by 1.85% (Fig. 3d).

**Tree cover changes in urbanized and urbanization areas**

To study the patterns of tree cover changes within newly urbanized areas, we used annual data on impervious surfaces to define in which year areas were urbanized (see Methods); that is converted into impervious surfaces (Fig. 4, Supplementary Fig. 9,10). Urbanized area built-up before 2000 have seen a moderate increase in tree cover by 4% from 2010 to 2019 (Fig. 4a), while areas being urbanized between 2006 and 2010 have a much higher increase in urban tree cover. For newly urbanized areas after 2010, lower increases or even decreases in tree cover were observed, likely because trees have not been planted or are still too small to be captured by the satellite system. New built-up areas after 2016 showed a loss of tree cover. These numbers overall suggest that tree planting and greening policies can balance the initial loss of trees in urbanization areas. At city-level, most large- and medium-sized cities that have experienced rapid urbanization after 2010 show an increase of urban tree cover within the urbanization areas (Fig. 4b). For example, Beijing shows an increase of 7.8% in urban tree cover in newly built-up areas (396 km$^2$) with the plantation of urban trees[4] (Supplementary Fig. 9). There is, however, also a number of small cities, such as Enshi (Hubei Province), which have experienced a net loss of tree cover in urban areas (-20%) without any greening (Supplementary Fig. 10).

Comparing greenness changes (reflected by MODIS NDVI) with tree cover changes between 2010 and 2019, we find a weak relationship ($r^2$=0.10), indicating the limited use of greenness as a proxy for urban tree cover changes (Fig. 4c).

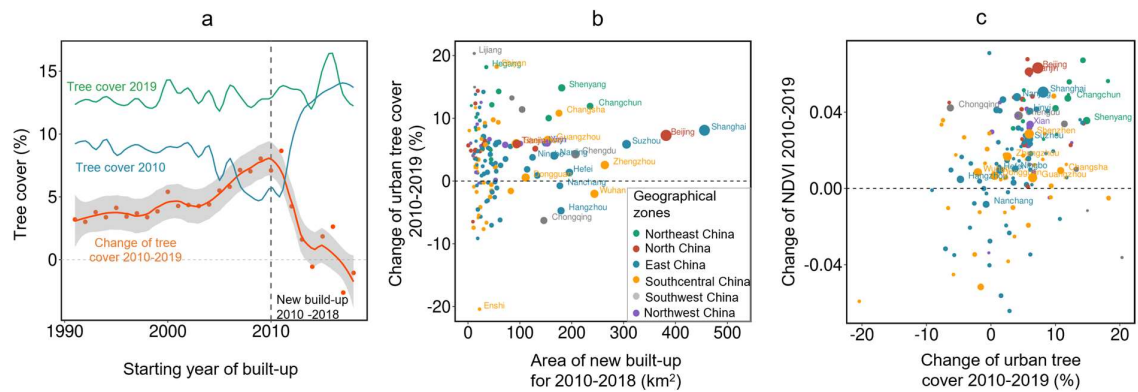

**Fig. 4: Tree cover changes and urbanization**. **a**, Mean tree cover for areas of different built-up time-steps during 1991-2018. **b**, Change in tree cover and urbanization areas for 2010-2019 at city level (n = 145). Colors denote geographical zones. **c**, Change in urban tree cover and yearly mean MODIS NDVI at city level (n = 145). Cities with urban areas ⩾ 500 km$^2$ were labeled with city names in b and c.

## Discussion

Urbanization in China promotes economic growth[33] and poverty reduction[34] but can at the same time cause environmental degradation, which challenges the sustainable development of

Chinese cities[35]. Urban trees are a key component of urban ecology and a possible pathway towards a higher life quality in large cities[36]. Consequently, the Chinese government has promoted the planting and maintenance of urban trees, aiming at mitigating the negative effects of urbanization and improve the urban environment[3].

Previous studies have shown a widespread greening of Chinese cities[2,18], but the somehow fuzzy variable termed "greenness" includes also grasses and shrubs etc., which do not provide the same level of ecological services as trees. Moreover, greenness maps are based on vegetation indices without units and are therefore less suited for quantifying changes, but are merely indicative of reporting directions of change. Our tree-level maps for 2010 and 2019 show a clear imprint of urban tree planting policies, but we also reveal that large cities, and in particular mega-cities, have a considerably higher tree cover and tree cover increase as compared to the majority of cities in China. This difference is likely related to differences in budgets allocated to urban tree plantations and management and reflects inequalities that are also observed at global scale[36]. Developed and wealthy regions, such as many cities in North America and Europe (or more generally the Global North), have made substantial investments in the planting and maintenance of urban trees[37–39], which arguably has improved the wellbeing of residents[17,40]. Contrastingly, many densely populated cities often located in the Global South, have limited resources for maintaining or increasing tree cover, which impacts people's health, for example via heatwaves[41,42], and higher depression risk[43]. These effects are aggravated by climatic conditions: in dryer regions, the costs of planting and managing of urban trees is higher, but at the same time the health benefits, such as the cooling effect, are more urgently needed[8].

While the high levels of tree cover increase in China's mega-cities seems to be a success-story at the first glance, the high maintenance costs[13] associated with irrigation[44] needs to be factored in when evaluating the sustainability of such projects. To reduce costs and ensure a sustainable development, a natural-based planning in early urbanization stages is required[45], including the use of local tree species that are adopted to the local climate and the preservation of existing trees. We found a considerable variability in tree cover among emerging cities, which is largely controlled by climatic conditions. In arid regions, it is advisable to select drought-tolerant tree species or implement alternative strategies such as incorporating short vegetation to ensure sustainable greening practices. Many fast-growing and cost-efficient tree species, such as willows and poplars[46], have been planted to deliver rapidly visible results, but the surviving rate is often low, and the low biodiversity[47] leads to increased vulnerability of trees to pests and diseases, which again increases **management costs. In response to these issues,** the Chinese government released new guidelines on the development of nature-based solutions for urban parks and forest in 2021, which emphasize the importance of selecting appropriate tree species and implementing rainwater harvesting for irrigation as important components of urban green policies[48].

Our study is based on commercial imagery, and the costs of repeated analyses at national level are currently not negligible for a large-sized country like China. However, the spatial resolution and coverage of publicly available data sources free of charge are not yet sufficient for mapping trees as single objects, often leaving a high uncertainty on mapping urban tree cover and in particular changes herein. Nevertheless, the costs of nanosatellite images are considerably lower as compared to traditional commercial sub-meter resolution imagery, and our study demonstrates that current technologies enable comprehensive monitoring of tree cover changes not only in Chinese cities but worldwide. This is expected to facilitate evidence-based decision-making and fostering global collaboration in urban greening initiatives for different countries as pledged by the UN Sustainable Development Goals 11$^{th}$ (Sustainable cities and communities) advocating for creating green public spaces[49].

# Methods

To calculate the change in urban tree canopy cover over the past decades, we define urban areas from a land cover map and selected the major 242 cities in China. We then map urban tree canopies using PlanetScope images from 2019 and RapidEye images from 2010 using a deep learning framework and compared the dynamics of urban trees between cities, as well as for urbanized and urbanization areas.

### Defining urban areas

We selected 242 cities by their size (area ⩾ 50 km$^2$), using the "artificial surface" class from the GlobaLand30 land cover map in 2020 at 30 m resolution. Areas classified as grassland and forest within built-up areas were included as urban areas. We also use Google Earth satellite imagery to double-check all urban boundaries, reviewing misclassifications and confirming the urban areas as spatial continuously built-up areas. Shijiazhuang city was omitted due to the lack of high-quality PlanetScope images for 2019. We then classified the cities into five groups according to the size of the urban area (Table 1, Supplementary Fig. 6), including 2 mega-cities (area ⩾ 1500 km$^2$), 4 large cities (area ⩾ 1,000 km$^2$), 19 medium cities (area ⩾ 500 km$^2$), 121 small cities (area ⩾ 100 km$^2$), and 96 emerging cities (area ⩾ 50 km$^2$) (Table 1, Supplementary Fig. 7). The cities were divided into six geographical zones to compare the regional differences in urban tree cover: northeast China (23 cities), north China (28 cities), east China (99 cities), south-central China (63 cities), northwest China (18 cities), and southwest China (11 cities) (Supplementary Fig. 7). RapidEye images covered 145 cities used to compare the change of urban tree cover between 2010 and 2019 (Table 1, Supplementary Fig. 7). The same urban boundaries from 2020 were used for both 2010 and 2019, implying that a given area may not have be urban in the 2010 map, and tree cover losses may therefore be related to transitions from forested land to build-up areas (assuming a dominant pattern of urban expansion and only rare cases of urban contraction[50]).

### Pre-processing PlanetScope and RapidEye

Here, we use PlanetScope images (4 bands: red, green, blue near-infrared) at 3 m spatial resolution to generate composites covering 242 cities in 2019. We organized and mosaicked raw satellite scenes in grids of 1×1 degree using images from periods where trees have full leaves, while grasses have passed their productivity peak[29]. We up-sampled Planet image from 3 m to 1 m using bilinear interpolation to preserve the high quality of the manual training samples and smooth the boundary of tree canopies[29].

The RapidEye images have a spatial resolution of 5 meters and are acquired in five spectral bands, including blue, green, red, red-edge, and near-infrared. We used RapidEye images from 2010, preprocessed in the same way as the PlanetScope images. Due to the lack of reliable meta-data on cloud cover, we only kept cloud-free RapidEye imagery for 145 cities by visually screening the images and disregarding cities with low data quality. Furthermore, a few patches within urban areas that had no available observations from either PlanetScope in 2019 or RapidEye in 2010 were also excluded from the analysis. This was done to ensure a consistent comparison of changes in tree cover within urban areas.

**Segmentation of tree canopies using deep learning**

We used the framework from Brandt et al[30] and Reiner et al[29] to segment tree canopy cover based on a convolutional neural network, specifically the U-Net architecture. We trained two models, one for PlanetScope and one for RapidEye. The models were trained with a batch size of 32 and a patch size of 256 x 256 pixels, and the Tversky loss was used as the loss function with $\alpha = 0.6$ and $\beta = 0.4$ to balance the commission and omission errors, see Supplementary Table 3 for specific settings. The training labels included individual tree crowns and clusters of trees and covered 209.29 km$^2$ over 496 sampled sites for 2019, including 34.74 km$^2$ of tree canopy cover distributed across 69 cities (Supplementary Fig.1a, Supplementary Table 3). For the 2010 Rapid-Eye data, we delineated tree canopies for 481 sites, covering 74.15 km$^2$ across 57 cities (Supplementary Fig.1b, Supplementary Table 3) and trained a model in the same way as for PlanetScope.

**Evaluation and comparison**

We compared our maps with an evaluation dataset consisting of 185 random 100m x 100m patches with manual labels from PlanetScope and RapidEye images, see Supplementary Fig. 10. The data used for evaluation was not used for training the models or selecting the hyper-parameters. The PlanetScope model showed an overall accuracy of 0.90, a kappa coefficient of 0.85, MSE of 0.37%, and RMSE of 7.69% (Supplementary Fig. 8a-c,f). The RapidEye model achieved an overall accuracy of 84%, a kappa coefficient of 0.78, MSE of -0.99%, and a RMSE of 9.84% (Supplementary Fig. 8a,b,d,g). To evaluate the uncertainty of the change between RapidEye and PlanetScope, we hand-labelled changes between 2010 and 2019 for the 185 patches and compared the results with the predictions. Here we obtained an $R^2$ of 0.83, a MSE of 1.07%, and an RMSE of 8.99% (Supplementary Fig. 8e), suggesting that the inter-comparison of tree cover maps derived from two different satellite systems is valid.

We compared our tree cover map in 2019 with other tree cover maps, including the MOD44B tree cover product and the ESA WorldCover 2020 tree cover map. Our map showed that MOD44B (spatial resolution of 250 m) underestimated tree cover in cities by 9.52% (Supplementary Fig. 11a). The mean urban tree cover of the Sentinel-2 based WorldCover map from 2020 was only 0.66% lower as compared to our map, but areas of low tree-cover were underestimated and areas of higher tree cover overestimated (Supplementary Fig. 11b). The results showed that 6.79% of the tree canopy cover in built-up areas was misclassified in the ESA WorldCover 2020 map (covering a total of 2182.12 km$^2$ of urban areas in our study) (Supplementary Table 2). The tree crowns omitted by the ESA map are often located in densely built-up areas, dominated by the existence of small and isolated trees. Additionally, almost half of the areas in the class "Tree cover" was found to be misclassified, as it was found to be dominated by shrubland, or grassland (Supplementary Table 2, Supplementary Fig. 6). The 2-m resolution tree cover map by ref.[51] and the Esri land cover map both underestimate urban tree cover, especially in the case of scattered trees (Supplementary Fig. 6).

**Development stages of built-up areas**

Annual maps of global artificial impervious surface areas (GAIA) for 1985-2018[50] at 30 m resolution were used to identify the starting year of built-up areas in 145 cities. The built-up areas continuously expanded over the past decades thereby indicating spatially explicitly the expansion of newly urbanized built-up areas (Supplementary Fig. 9c,10c). The different starting years where built-up were first observed serve as an indicator of the development stage of urbanized and urbanization areas. Built-up areas starting after 2010 indicate here urbanization areas for 2010-2018, while built-up areas constructed before 2010 were regarded as urbanized areas. The mean tree cover in 2010 and 2019 for annual new built-up areas was studied to compare the tree cover change in built-up areas at various development stages.

**Population data**

We used the WorldPop dataset for 2019 to quantify the population density for each city. WorldPop provides the estimated number of people residing in a 100 × 100 m grid based on a random forest model and a global dataset including administrative unit-based census information, which has a higher spatial resolution and update frequency than other population datasets[52].

**Urban greening**

To analyze the dynamics of vegetation greenness in 145 cities, we combined the MODIS Collection 6 Vegetation Index (VI) product from MOD13Q1 C6 and MYD13Q1 C6. To filter out observations affected by snow, water, and cloud cover, we utilized the embedded quality data (SummaryQA ≤ 1) available in Google Earth Engine. The maximum monthly NDVI (Normalized Difference Vegetation Index) was calculated, and any no-data pixels were filled using the mean values from a four-month moving window (including the previous two months and following two months). Annual average values were then derived from the monthly NDVI time series for the period 2001-2020.

**Code availability**

The code for the tree canopy detection framework based on U-Net is publicly available at https://doi.org/10.5281/zenodo.3978185.

**Data and materials availability**

PlanetScope imagery and RapidEye imagery in urban areas over China is available from Planet Labs (https://www.planet.com/products/) upon acquiring a license agreement. GlobaLand30 land cover dataset is available at http://www.globallandcover.com/home_en.html. ESA WorldCover 2020 land cover map can be downloaded at https://worldcover2020.esa.int/. Annual maps for the global artificial impervious areas (GAIA) dataset can be downloaded from http://data.ess.tsinghua.edu.cn. MODIS NDVI products, including MOD13Q1 (https://lpdaac.usgs.gov/products/mod13q1v061/) and MCD12Q2 (https://lpdaac.usgs.gov/products/mcd12q2v006/), are available from the google earth engine (https://earthengine.google.com). The administrative boundary in China is accessible from the national catalogue service for geographic information (https://www.ngcc.cn/).

## Acknowledgments

X.Z. was funded by the China Scholarship Council (CSC, grant no. 20190). M.B. was funded by the European Research Council (ERC) under the European Union's Horizon 2020 Research and Innovation Programme (grant agreement no. 947757 TOFDRY) and DFF Sapere Aude (grant no. 9064–00049B). X.T was funded by the National Natural Science Foundation of China for Excellent Young Scientists (Overseas). F.T. acknowledges funding from the Nation Natural Science Foundation of China (Grant No. 42001299) and the Seed Fund Program for Sino-Foreign Joint Scientific Research Platform of Wuhan University (No. WHUZZJJ202205). Y.Y. was funded by International Partnership Program of Chinese Academy of Sciences (092GJHZ2022029GC) and the CAS Interdisciplinary Innovation Team (JCTD-2021-16). R.F. acknowledges support from Villum Fonden through the project Deep Learning and Remote Sensing for Unlocking Global Ecosystem Resource Dynamics (DeReEco).

## Contributions

X.Z. and M.B. designed the research. F.R. and X.T. developed the code for the collection of PlanetScope and RapidEye images, and F.R. and S.L. developed the code for the deep learning pipeline. X.Z. prepared the annotation data and conducted the experiments and analysis. X.Z., M.B. and R.F. drafted the first manuscript and all authors contributed the discussion and the final version of the manuscript.

## Competing interest

The authors declare no competing interests.

## References

1. Liu, Y. *et al.* Correlations between Urbanization and Vegetation Degradation across the World's Metropolises Using DMSP/OLS Nighttime Light Data. *Remote Sens.* **7**, 2067–2088 (2015).

2. Zhang, X. *et al.* A large but transient carbon sink from urbanization and rural depopulation in China. *Nat. Sustain.* **5**, 321–328 (2022).
3. Feng, D., Bao, W., Yang, Y. & Fu, M. How do government policies promote greening? Evidence from China. *Land Use Policy* **104**, (2021).
4. Yao, N. *et al.* Beijing's 50 million new urban trees: Strategic governance for large-scale urban afforestation. *Urban For. Urban Green.* **44**, 126392 (2019).
5. Liu, C. *et al.* Urban forestry in China: status and prospects. *Urban Agric. Mag.* **13**, 15–17 (2004).
6. Jackson, L. E. The relationship of urban design to human health and condition. *Landsc. Urban Plan.* **64**, 191–200 (2003).
7. Endreny, T. A. Strategically growing the urban forest will improve our world. *Nat. Commun.* **9**, 1160 (2018).
8. Manoli, G. *et al.* Magnitude of urban heat islands largely explained by climate and population. *Nature* **573**, 55–60 (2019).
9. Schwaab, J. *et al.* The role of urban trees in reducing land surface temperatures in European cities. *Nat. Commun.* **12**, 6763 (2021).
10. Iungman, T. *et al.* Cooling cities through urban green infrastructure: a health impact assessment of European cities. *The Lancet* **401**, 577–589 (2023).
11. Ko, Y. Trees and vegetation for residential energy conservation: A critical review for evidence-based urban greening in North America. *Urban For. Urban Green.* **34**, 318–335 (2018).
12. Locosselli, G. M. *et al.* The role of air pollution and climate on the growth of urban trees. *Sci. Total Environ.* **666**, 652–661 (2019).
13. Zhang, B., Xie, G., Zhang, C. & Zhang, J. The economic benefits of rainwater-runoff reduction by urban green spaces: A case study in Beijing, China. *J. Environ. Manage.* **100**, 65–71 (2012).
14. Nowak, D. J. & Crane, D. E. Carbon storage and sequestration by urban trees in the USA. *Environ. Pollut.* **116**, 381–389 (2002).
15. Alvey, A. A. Promoting and preserving biodiversity in the urban forest. *Urban For. Urban Green.* **5**, 195–201 (2006).
16. Sander, H., Polasky, S. & Haight, R. G. The value of urban tree cover: A hedonic property price model in Ramsey and Dakota Counties, Minnesota, USA. *Ecol. Econ.* **69**, 1646–1656 (2010).
17. Nowak, D. J., Hirabayashi, S., Bodine, A. & Greenfield, E. Tree and forest effects on air quality and human health in the United States. *Environ. Pollut.* **193**, 119–129 (2014).
18. Liu, X. *et al.* High-spatiotemporal-resolution mapping of global urban change from 1985 to 2015. *Nat. Sustain.* (2020) doi:10.1038/s41893-020-0521-x.
19. Zhang, L. *et al.* Direct and indirect impacts of urbanization on vegetation growth across the world's cities. *Sci. Adv.* **8**, eabo0095 (2022).
20. Zhao, S., Liu, S. & Zhou, D. Prevalent vegetation growth enhancement in urban environment. *Proc. Natl. Acad. Sci.* **113**, 6313 (2016).
21. Shahtahmassebi, A. R. *et al.* Remote sensing of urban green spaces: A review. *Urban For. Urban Green.* **57**, 126946 (2021).
22. Pu, R. & Landry, S. A comparative analysis of high spatial resolution IKONOS and WorldView-2 imagery for mapping urban tree species. *Remote Sens. Environ.* **124**, 516–533 (2012).
23. Neyns, R. & Canters, F. Mapping of Urban Vegetation with High-Resolution Remote Sensing: A Review. *Remote Sens.* **14**, 1031 (2022).
24. Pu, R. & Landry, S. Mapping urban tree species by integrating multi-seasonal high resolution pléiades satellite imagery with airborne LiDAR data. *Urban For. Urban Green.* **53**, 126675 (2020).
25. Yadav, S., Rizvi, I. & Kadam, S. Urban tree canopy detection using object-based image analysis for very high resolution satellite images: A literature review. in *2015 International Conference on Technologies for Sustainable Development (ICTSD)* 1–6 (2015). doi:10.1109/ICTSD.2015.7095889.
26. Erker, T., Wang, L., Lorentz, L., Stoltman, A. & Townsend, P. A. A statewide urban tree canopy mapping method. *Remote Sens. Environ.* **229**, 148–158 (2019).
27. Guo, J., Xu, Q., Zeng, Y., Liu, Z. & Zhu, X. X. Nationwide urban tree canopy mapping and coverage assessment in Brazil from high-resolution remote sensing images using deep learning. *ISPRS J. Photogramm. Remote Sens.* **198**, 1–15 (2023).
28. Ma, Q. *et al.* Individual structure mapping over six million trees for New York City USA. *Sci. Data* **10**, 102 (2023).
29. Reiner, F. *et al.* More than one quarter of Africa's tree cover is found outside areas previously classified as forest. *Nat. Commun.* **14**, 2258 (2023).
30. Brandt, M. *et al.* An unexpectedly large count of trees in the West African Sahara and Sahel. *Nature* **587**, 78–82 (2020).

31. Pu, R., Landry, S. & Yu, Q. Assessing the potential of multi-seasonal high resolution Pléiades satellite imagery for mapping urban tree species. *Int. J. Appl. Earth Obs. Geoinformation* **71**, 144–158 (2018).
32. Zanaga, D. *et al.* ESA WorldCover 10 m 2020 v100. (2021) doi:10.5281/zenodo.5571936.
33. Chen, M., Zhang, H., Liu, W. & Zhang, W. The global pattern of urbanization and economic growth: evidence from the last three decades. *PloS One* **9**, e103799 (2014).
34. Zhang, Y. Urbanization, Inequality, and Poverty in the People's Republic of China.
35. He, C., Gao, B., Huang, Q., Ma, Q. & Dou, Y. Environmental degradation in the urban areas of China: Evidence from multi-source remote sensing data. *Remote Sens. Environ.* **193**, 65–75 (2017).
36. Roloff, A. *Urban tree management: for the sustainable development of green cities*. (John Wiley & Sons, 2016).
37. Kroeger, T., McDonald, R. I., Boucher, T., Zhang, P. & Wang, L. Where the people are: Current trends and future potential targeted investments in urban trees for PM10 and temperature mitigation in 27 US cities. *Landsc. Urban Plan.* **177**, 227–240 (2018).
38. Wild, T., Freitas, T. & Vandewoestijne, S. *Nature-based solutions: State of the art in EU-funded projects*. (Publications Office of the European Union, 2020).
39. Eisenman, T. S., Flanders, T., Harper, R. W., Hauer, R. J. & Lieberknecht, K. Traits of a bloom: a nationwide survey of US urban tree planting initiatives (TPIs). *Urban For. Urban Green.* **61**, 127006 (2021).
40. Morani, A., Nowak, D. J., Hirabayashi, S. & Calfapietra, C. How to select the best tree planting locations to enhance air pollution removal in the MillionTreesNYC initiative. *Environ. Pollut.* **159**, 1040–1047 (2011).
41. Wei, H., Chen, B., Wu, S. & Xu, B. Impact of early heat anomalies on urban tree cooling efficiency: Evidence from spring heatwave events in India. *Int. J. Appl. Earth Obs. Geoinformation* **120**, 103334 (2023).
42. Yang, J. *et al.* Heatwave and mortality in 31 major Chinese cities: definition, vulnerability and implications. *Sci. Total Environ.* **649**, 695–702 (2019).
43. Chen, T.-H. K. *et al.* Higher depression risks in medium- than in high-density urban form across Denmark. *Sci. Adv.* **9**, eadf3760 (2023).
44. Pan, L., Zhao, Y. & Zhu, T. Estimating Urban Green Space Irrigation for 286 Cities in China: Implications for Urban Land Use and Water Management. *Sustainability* **15**, 8379 (2023).
45. Van den Bosch, M. & Sang, Å. O. Urban natural environments as nature-based solutions for improved public health–A systematic review of reviews. *Environ. Res.* **158**, 373–384 (2017).
46. Li, X. *et al.* Developing a sub-meter phenological spectral feature for mapping poplars and willows in urban environment. *ISPRS J. Photogramm. Remote Sens.* **193**, 77–89 (2022).
47. Su, Z., Zhang, R. & Qiu, J. Decline in the diversity of willow trunk-dwelling weevils (Coleoptera: Curculionoidea) as a result of urban expansion in Beijing, China. *J. Insect Conserv.* **15**, 367–377 (2011).
48. Qi, J. J. & Dauvergne, P. China and the global politics of nature-based solutions. *Environ. Sci. Policy* **137**, 1–11 (2022).
49. Cf, O. Transforming our world: the 2030 Agenda for Sustainable Development. *U. N. N. Y. NY USA* (2015).
50. Gong, P. *et al.* Annual maps of global artificial impervious area (GAIA) between 1985 and 2018. *Remote Sens. Environ.* **236**, 111510 (2020).
51. He, D., Shi, Q., Liu, X., Zhong, Y. & Zhang, L. Generating 2m fine-scale urban tree cover product over 34 metropolises in China based on deep context-aware sub-pixel mapping network. *Int. J. Appl. Earth Obs. Geoinformation* **106**, 102667 (2022).
52. Tatem, A. J. WorldPop, open data for spatial demography. *Sci. Data* **4**, 1–4 (2017).

# Supplementary Material

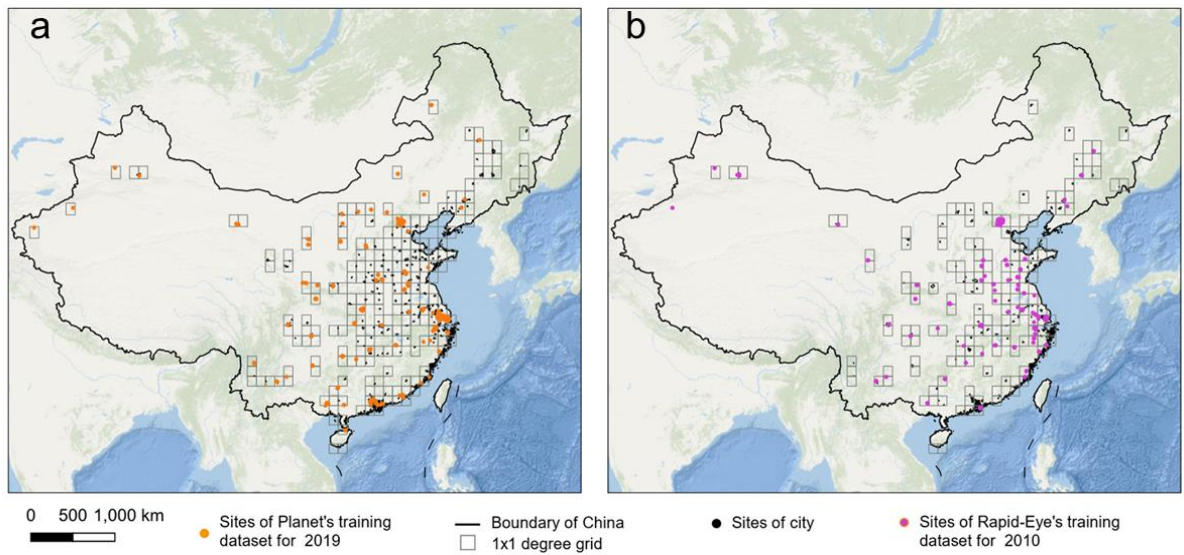

**Supplementary Fig. 1. Sites of training-label parcels for the deep-learning model**. **a**, Training parcels for PlanetScope imagery in 2019; **b**, Training parcels for RapidEye imagery in 2010. The Ocean Basemap is provided by Esri.

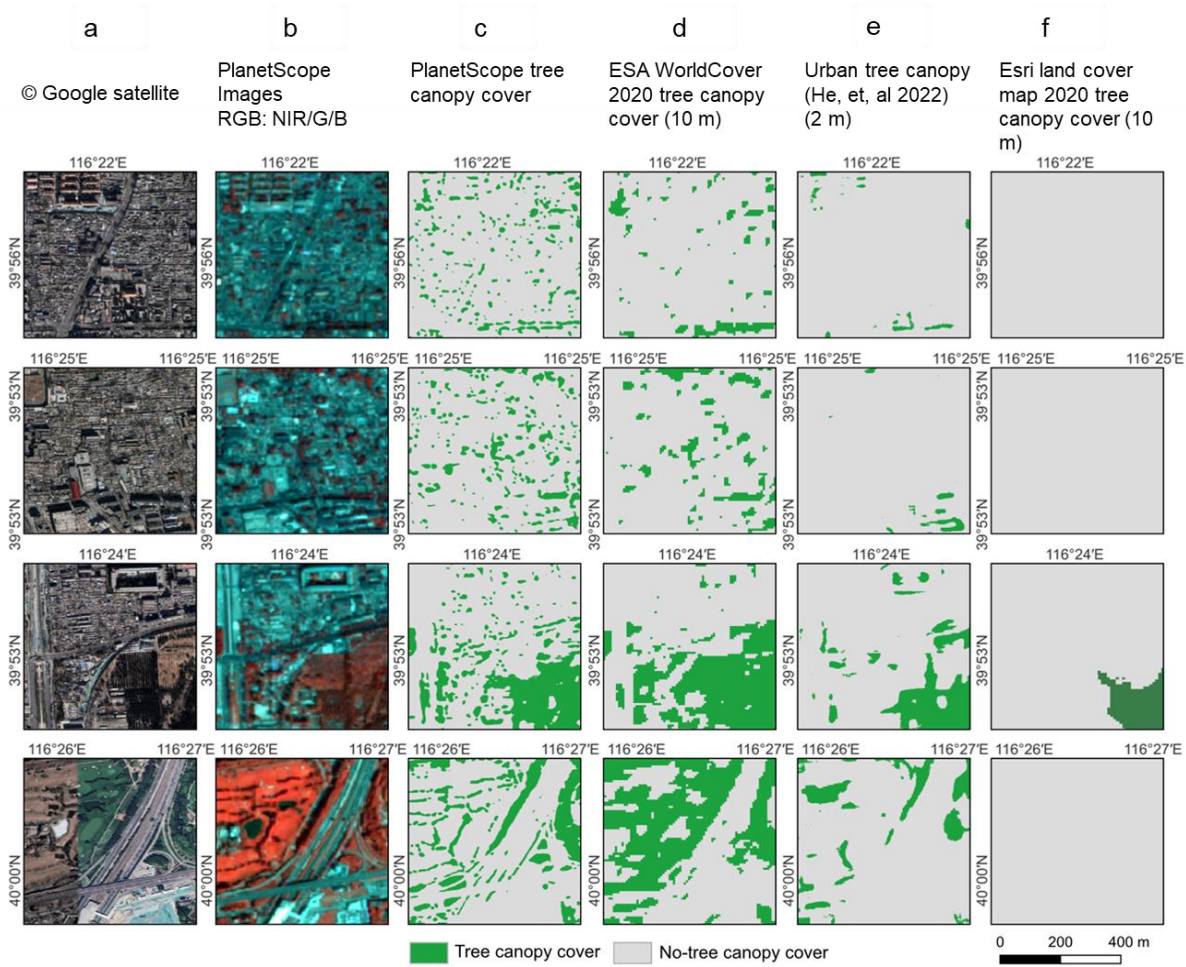

**Supplementary Fig. 2. Comparison of PlanetScope tree canopy mapping with other products**. **a**, Google Earth satellite images. **b**, PlanetScope Image 2019 (RGB: NIR/G/B). **c**, PlanetScope tree canopy mapping 2019. **d**, Tree canopy from ESA 2020 Land cover map[1]. **e**, 2 m fine-scale urban tree canopy map from He et, al 2022[2]. **f**, Tree canopy based on Esri land cover map 2020[3].

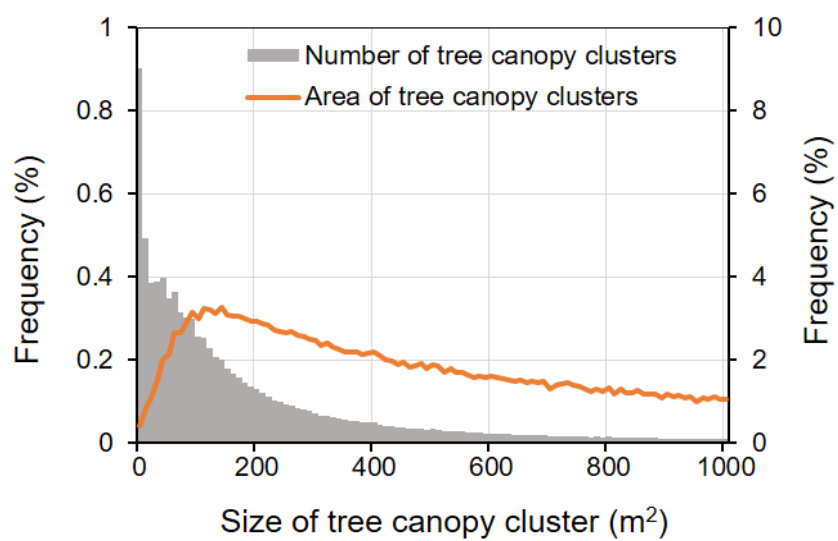

**Supplementary Fig. 3. Histogram of the size of tree canopy clusters in urban areas.**

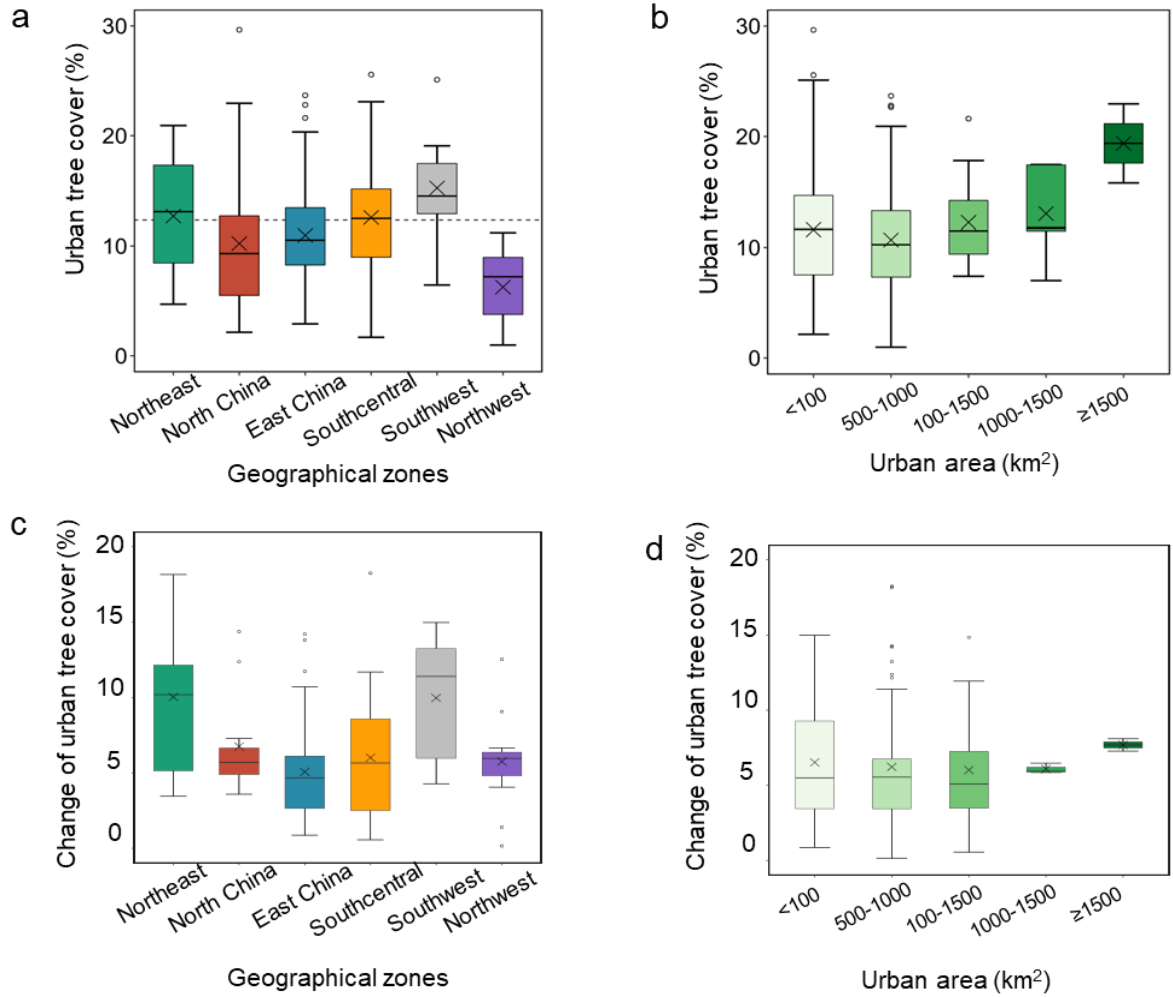

**Supplementary Fig. 4. Urban tree cover 2019 and change from 2010 to 2019 at city level.**
**a**, Urban tree cover in 2019 grouped by city area, and (**b**) by geographical zones. **c**, Urban tree cover change from 2010 to 2019 at city level grouped by city size, and (**d**) by geographical zones.

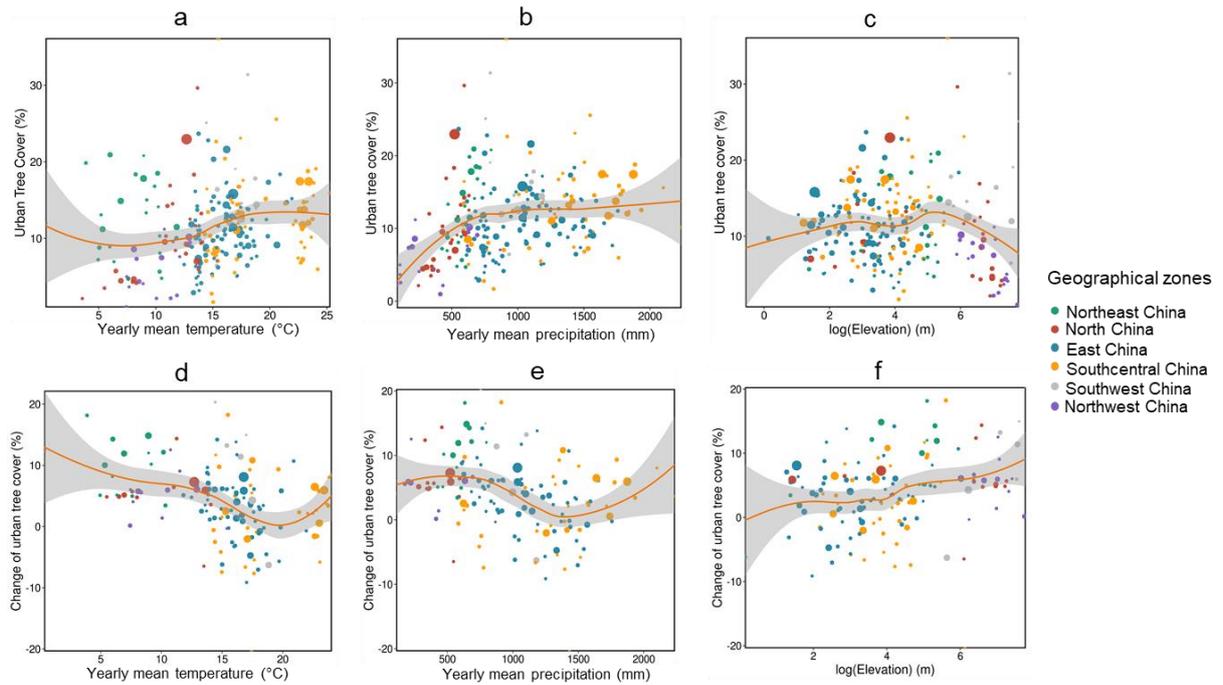

**Supplementary Fig. 5. Urban tree cover for 2019 as well as change in urban tree cover from 2010 to 2019 in relation to climate/elevation variables**. Mean annual temperature in 2015 (**a**), mean precipitation in 2015 (**b**) and elevation (**c**) at city level (n = 242). Change in urban tree cover during 2010-2019 in relation to mean annual temperature in 2015 (**d**) and mean precipitation in 2015 (**e**) and elevation (**f**) at city level (n =145). The fitted orange lines and confidence areas were fitted by a loess function in R.

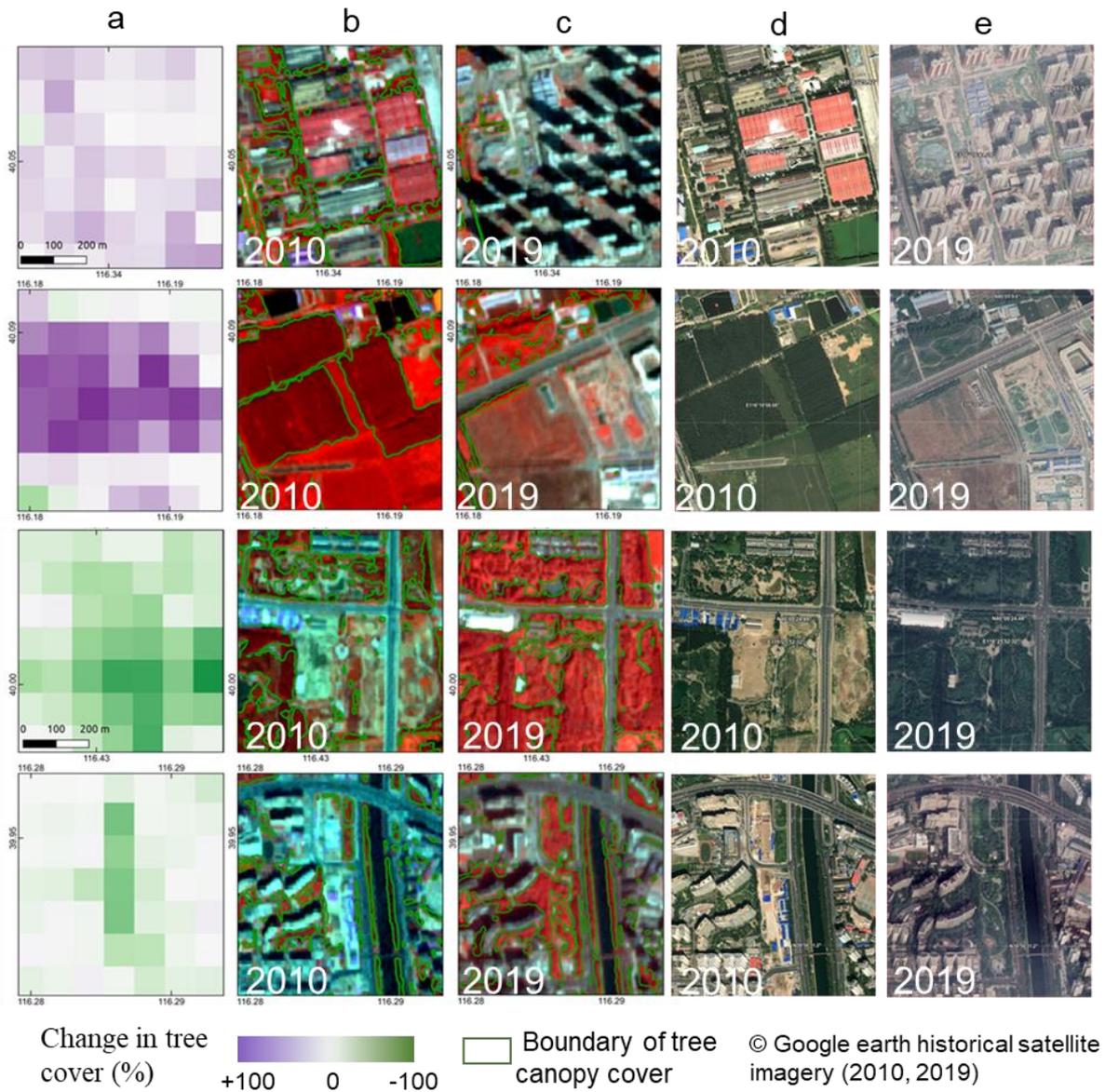

**Supplementary Fig. 6. Examples showing an increase in tree canopy cover from 2010 to 2019. a**, Change in tree cover in 1 ha grids (2010 - 2019). **b**, Prediction of tree canopy cover based on RapidEye imagery for 2010. c, Prediction of tree canopy cover based on PlanetScope imagery for 2019. **d**, Google Earth historical imagery in 2010. **e**, Google Earth historical imagery in 2019.

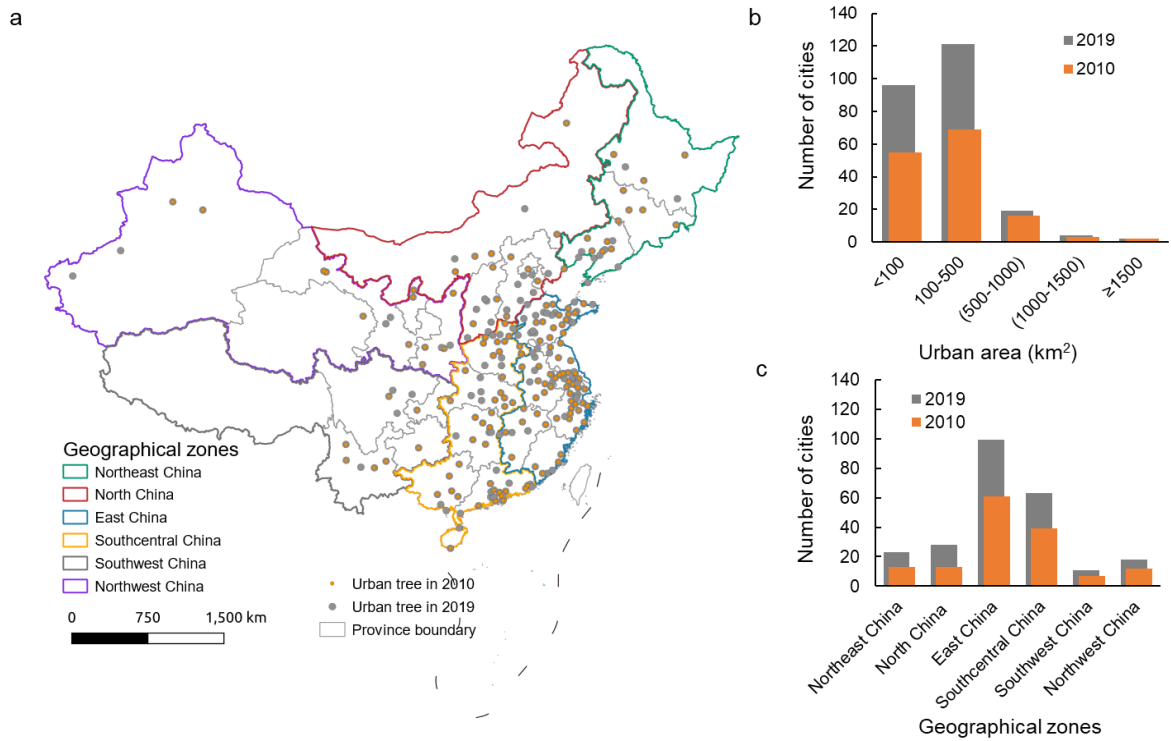

**Supplementary Fig. 7. Cities studied in 2010 and 2019**. **a**, Spatial distribution of cities studied in 2010 and 2019. **b**, Size of cities analyzed. **c**, Number of cities in the different geographical zones.

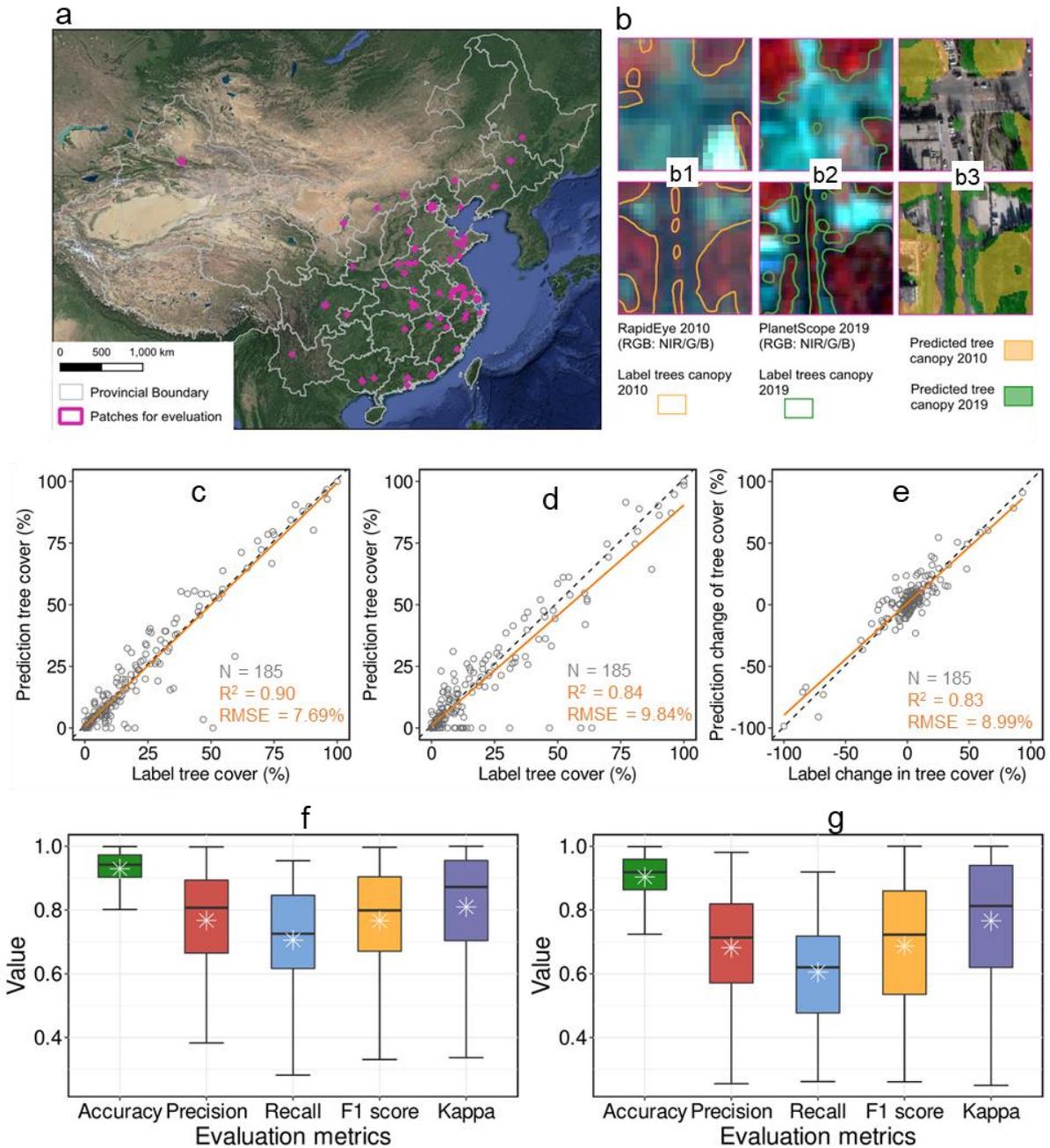

**Supplementary Fig. 8. Comparison between manually labeled areas from the test dataset and the corresponding predictions for 185 patches (the size of each patch is 1 ha)**. **a**, Location of patches for evaluation. **b**, examples of patches with labelled tree canopy cover for 2010 (b1) and 2019 (b2) and prediction (b3). **c**, Comparison between predictions and manual labelling for PlanetScope 2019 tree canopy cover. **d**, Comparison between predictions and manual labelling for RapidEye 2010 tree canopy cover. **e**, Comparison of tree canopy cover changes from 2010 to 2019 between model predictions and manual labelling. **f**, Statistical evaluation metrics for the PlanetScope 2019 tree canopy cover mapping. **g**, Statistical evaluation metrics for the RapidEye 2010 tree canopy cover mapping.

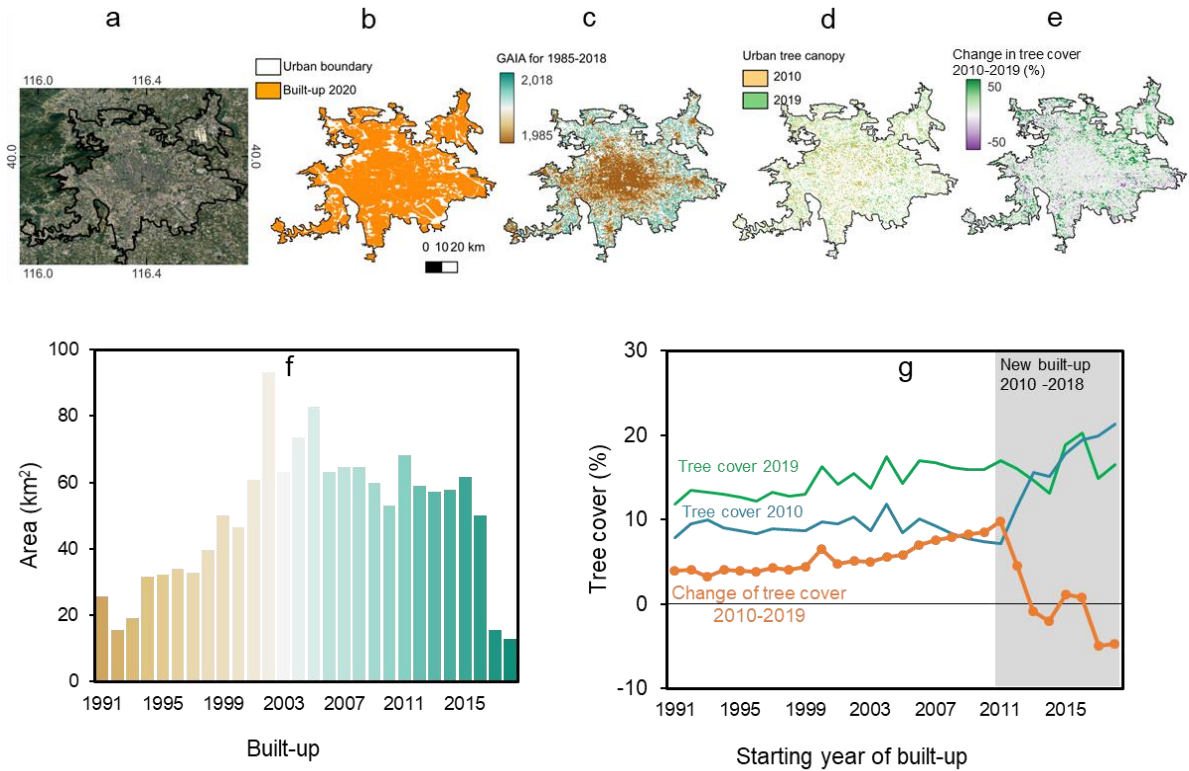

**Supplementary Fig. 9. Change in tree cover and urbanization in Beijing**. **a**, Google satellite image. **b**, Built-up areas in 2020 and the urban boundary from the GlobalLand30 2020 map. **c**, Expansion of built-up areas from GAIA 1991-2018[4]. **d**, Tree canopy cover in 2010 and 2019. **e**, Change of tree cover between 2010 and 2019 in 1-ha grids. **f**, Area of new built-up from 1991-2018. **g**, Mean tree cover for areas of different built-up time-steps during 1991-2018.

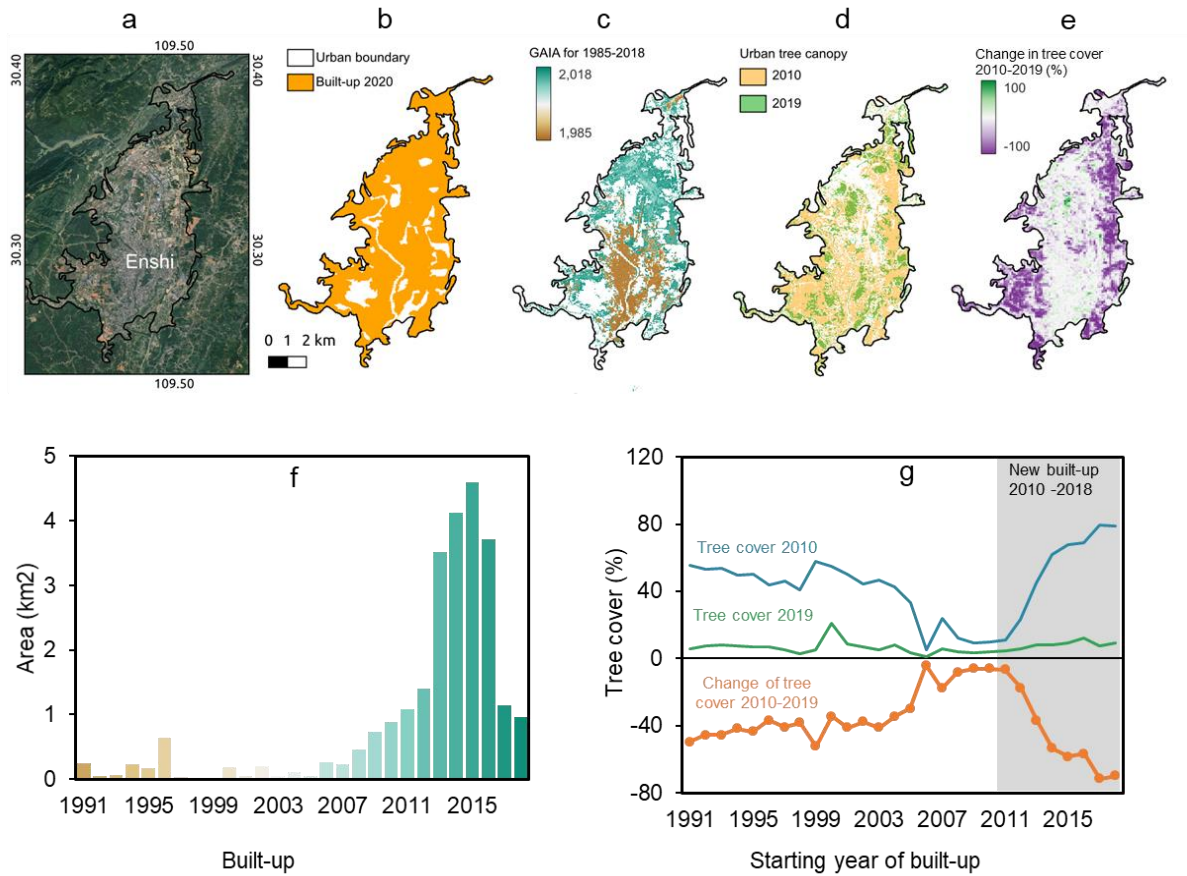

**Supplementary Fig. 10. Change in tree cover and urbanization in Enshi**. **a**, Google satellite image. **b**, Built-up areas in 2020 and the urban boundary from the GlobalLand30 2020 map. **c**, Expansion of built-up areas from GAIA 1991-2018[4]. **d**, Tree canopy cover in 2010 and 2019. **e**, Change in tree cover between 2010 and 2019 in 1-ha grids. **f**, Area of new built-up from 1991-2018. **g**, Mean tree cover for areas of different built-up time-steps during 1991-2018.

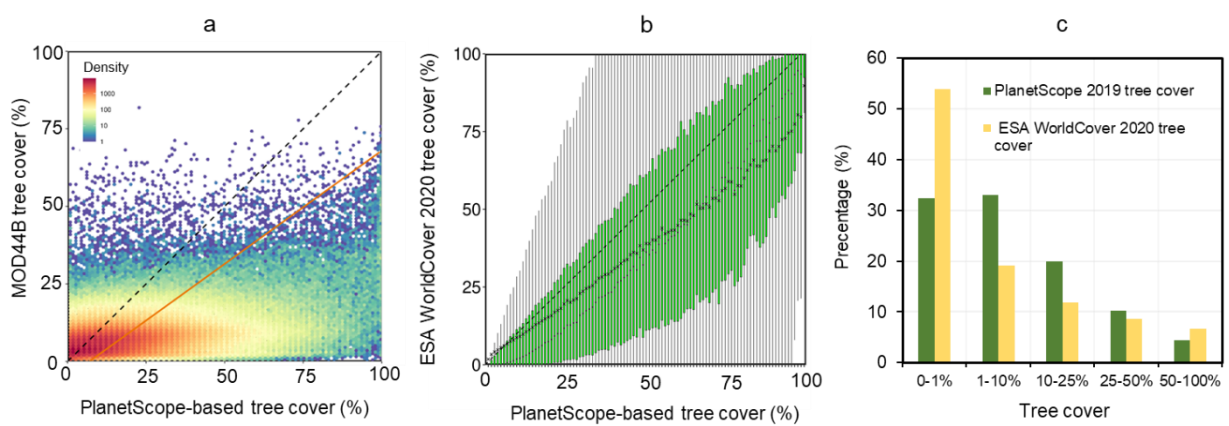

**Supplementary Fig. 11. Comparison of tree cover predictions from PlanetScope estimations and other tree cover products. a**, Density plot for the PlanetScope-based 2019 tree cover and MOD44B 2019 tree cover. **b**, Density plot for the PlanetScope-based 2019 tree cover and ESA WorldCover 2020 tree cover. **c**, histogram of Planet 2019 tree cover and ESA WorldCover 2020 tree cover.

**Supplementary Table 1. Datasets used in the study.**

| Dataset | Resolution | Bands | Count of cities | Areas (km²) |
|---|---|---|---|---|
| **PlanetScope Imagery 2019** | ~3 m | Blue, Green, Red, NIR | 242 | 51882 |
| **RapidEye Imagery 2010** | ~5 m | Blue, Green, Red, Red Edge, NIR | 145 | 36068 |
| **GAIA 2000 - 2018** | 30 m | Impervious surface | 242 | 51882 |
| **MODIS vegetation index (MOD13Q1 C6, MYD13Q1 C6)** | 250 m | NDVI | 242 | 51882 |

**Supplementary Table 2. Mean tree cover by land cover class. Land cover classes are derived from the ESA WorldCover 2020 map[1].**

| Land cover type (ESA WorldCover 2020) | Total area (km²) | PlanetScope tree cover (%) |
|---|---|---|
| Tree cover | 65291.74 | 48.26 |
| Shrubland | 806.22 | 23.02 |
| Grassland | 11301.15 | 14.24 |
| Cropland | 74083.89 | 17.59 |
| Built-up | 401606.23 | 6.79 |
| Bare/sparse vegetation | 85634.48 | 6.41 |
| Permanent water bodies | 9461.42 | 5.27 |
| Herbaceous wetland | 161.62 | 12.46 |

**Supplementary Table 3. Information of samples fed into the U-Net model and the core model settings for the analysis.**

| PlanetScope 2019 | | RapidEye 2010 | | Training settings | |
|---|---|---|---|---|---|
| **Inputs** | **Count** | **Inputs** | **Count** | **Hyperparameters** | **Setting** |
| Box with positive and negative samples | 418 | Box with positive and negative samples | 423 | Sampling Strategy | Sequential patches with 1/4 overlap |
| Box with only negative samples | 78 | Box with only negative samples | 58 | Patch Size | 256 |
| Box | 496 | Box | 481 | Batch Size | 8 |
| Positive Samples (km$^2$) | 34.74 | Positive Samples (km$^2$) | 12.89 | Training Steps | 100 |
| Negative Samples (km$^2$) | 174.55 | Negative Samples (km$^2$) | 67.88 | Number of Epochs | 500 |
| Distributed cities | 69 | Distributed cities | 57 | Ratio in Tversky Loss | 0.6, 0.4 |
| Bands | 4 (Blue, Green, Red, NIR) | Bands | 5 (Blue, Green, Red, Red Edge, NIR) | Resampled-scale | 3 |

## Reference

1. Zanaga, D. *et al.* ESA WorldCover 10 m 2020 v100. (2021) doi:10.5281/zenodo.5571936.
2. He, D., Shi, Q., Liu, X., Zhong, Y. & Zhang, L. Generating 2m fine-scale urban tree cover product over 34 metropolises in China based on deep context-aware sub-pixel mapping network. *Int. J. Appl. Earth Obs. Geoinformation* **106**, 102667 (2022).
3. Karra, K. *et al.* Global land use/land cover with Sentinel 2 and deep learning. in 4704–4707 (IEEE, 2021).
4. Gong, P. *et al.* Annual maps of global artificial impervious area (GAIA) between 1985 and 2018. *Remote Sens. Environ.* **236**, 111510 (2020).